# TransfQMix: Transformers for Leveraging the Graph Structure of Multi-Agent Reinforcement Learning Problems

Matteo Gallici
KEMLG Research Group, Universitat
Politècnica de Catalunya
Barcelona, Spain
gallici@cs.upc.edu

Mario Martin
KEMLG Research Group, Universitat
Politècnica de Catalunya
Barcelona, Spain
mmartin@cs.upc.edu

Ivan Masmitja
Institut de Ciències del Mar (ICM),
CSIC
Barcelona, Spain
masmitja@icm.csic.es

## ABSTRACT

Coordination is one of the most difficult aspects of multi-agent reinforcement learning (MARL). One reason is that agents normally choose their actions independently of one another. In order to see coordination strategies emerging from the combination of independent policies, the recent research has focused on the use of a centralized function (CF) that learns each agent's contribution to the team reward. However, the structure in which the environment is presented to the agents and to the CF is typically overlooked. We have observed that the features used to describe the coordination problem can be represented as vertex features of a latent graph structure. Here, we present TransfQMix, a new approach that uses transformers to leverage this latent structure and learn better coordination policies. Our transformer agents perform a graph reasoning over the state of the observable entities. Our transformer Q-mixer learns a monotonic mixing-function from a larger graph that includes the internal and external states of the agents. TransfQMix is designed to be entirely transferable, meaning that same parameters can be used to control and train larger or smaller teams of agents. This enables to deploy promising approaches to save training time and derive general policies in MARL, such as transfer learning, zero-shot transfer, and curriculum learning. We report TransfQMix's performances in the Spread and StarCraft II environments. In both settings, it outperforms state-of-the-art Q-Learning models, and it demonstrates effectiveness in solving problems that other methods can not solve.

## 1 INTRODUCTION

In order to solve cooperative multi-agent problems, it is critical that agents behave in a coordinated manner. Deep reinforcement learning (RL) has been successfully applied to numerous multi-agent optimization tasks [6, 8, 12]. When we try to apply RL to learn coordination policies, however, we face numerous challenges. Due to communication constraints, the deployment of a central controller is not practical. Even when communication is allowed, the large size of the observation and action spaces introduces the curse of dimensionality, discouraging the use of a single actuator. Agents should therefore choose their actions independently of one another. In order to see coordinating strategies emerging from the combination of independent policies, state-of-the-art multi-agent reinforcement learning (MARL) models use one or more centralized functions (CFs) to learn the contribution of the agents' actions to the team goal. The CFs allow to optimize the agents' parameters with respect to a global team reward. Once trained, they can still be deployed autonomously since each agent is in charge of choosing its own behavior. This approach is referred to as the *centralized-training-decentralized-execution* (CTDE) paradigm [4, 7].

During the last years, most of the works have focused on the CFs of CTDE. Methods such as Value Decomposition Networks (VDN) [21], QMix [18], and QTran [20] extended the traditional Q-Learning algorithm [28] with a central network that (learns to) project the agent's action-values over the q-value of the joint action. Actor-critic models such as Multi-Agent Deep Deterministic Policy gradient (MADDPG) [11] and Multi-Agent Proximal Policy Optimization (MAPPO) [30], allow the critic networks to access global observations during training. More recent approaches, like Deep Implicit Coordination Graphs (DICG) [9] and QPlex [26] refined the CF with the use of multi-head self-attention and graph neural networks. Nonetheless, individual agents are usually kept simple by employing recurrent neural networks (RNN) fed by observation vectors that are large concatenations of various types of features (see Figure 1a). By performing these concatenations a key information is lost: the fact that many of the features are exactly of the same type despite referring to separate entities (e.g., the position in a map).

Our work shows that the structure of the observation space, as well as the architecture used to deploy the agents and the CFs, play an important role in solving complex coordination tasks. We suggest that observation vectors contain mostly vertex features of a latent graph structure that becomes explicit when we reconsider how they are fed into neural networks. Consequently, instead of chaining together many features to generate a vector that describes the state of the *world* observed by the agent, we generalize a set of features and we use them to describe the state of the *entities* observed by the agent (or the CFs). Our approach is depicted in Figure 1b. We do not include any additional information in this process. On the contrary, sometimes we need to remove data that is not accessible for all the observed entities. There are several advantages in this approach: (i) we can employ the same weights of an embedded feed forward network to process the same vertex features, reducing the complexity of the feature space; and (ii), we can learn the edges of the latent coordination graph using a self-attention mechanism. In particular, we employ transformers [24], which have been shown to be an effective graph-based architecture in natural language processing [27], computer vision [29], and even for developing a generalist agent [19].

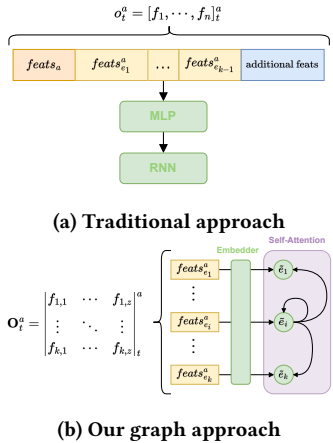

(a) Traditional approach

(b) Our graph approach

Figure 1: A traditional observation vector and our graph approach. In traditional approaches (a), the observation vector for the agent $a$ at the time step $t$ is defined by a concatenation of features relative to itself, to the other $k-1$ entities, and to additional elements (e.g., previous actions). In our approach (b), we keep only the $z$ features defined for all the entities to generate the vertices of a coordination graph, the edges of which are learned via a self-attention mechanism.

Our transformer agents sample their actions after processing the graph of entities observed at a specific time step. Our transformer Q-mixer learns a monotonic mixing-function from a larger graph that contains the agents' internal and external states. Given the strong temporal dependencies in RL problems, we add a recurrent mechanism in both the agents and the mixer, which allows us to affect the graph reasoning at a certain time step with an embedding of the preceding. The resulting model, TransfQMix, has the advantage of being totally transferable, meaning that the same parameters can be applied to control and train larger or smaller teams. This is allowed since the networks' weights constitute an attention mechanism that is independent of the number of vertices to which it is applied. Traditional models, conversely, must be re-trained every time we introduce a new entity, because the dimension of the concatenated vectors changes, and therefore the networks weights must be readjusted. The total transferability of TransfQMix enables to deploy transfer learning, zero-shot transfer, and curriculum learning, which are crucial steps towards more general models in MARL.

We tested TransfQMix in multiple scenarios of the Spread task [11], and in the hardest maps of StarCraft II (SC2) [25]. TransfQMix outperformed state-of-the-art Q-Learning models in both environments, and it could solve problems that others can not address, showing in general faster convergence to better coordination policies.

The following is a list of the contributions of this paper:

(1) We formalize a new paradigm for cooperative MARL, which consists of rethinking the coordination tasks as graph embedding tasks.
(2) We present a new method, TransfQMix, that uses transformers to leverage coordination graphs and outperforms state-of-the-art Q-Learning methods.
(3) We introduce a graph-based recurrent mechanism for including a time dependency in both the transformer agents and mixer.
(4) We design TransfQMix to be able to process graphs of entities of varying sizes. This allows us to obtain a more general method which can be used to deploy zero-shot transfer, transfer learning, and curriculum learning in MARL.

## 2 RELATED WORK

Recent state-of-the-art methods tackle MARL problems using the CTDE paradigm [4, 7]. The CTDE approach was deployed successfully with policy-based and value-based methods [9, 30, 31]. Here, we have focused on value-based methods that use CTDE.

A necessary condition for implementing CTDE effectively in multi-agent Q-Learning is that a greedy sampling of the joint action is equivalent to sampling the actions greedily from the individual agents [25]. This principle is known as the individual-global-max (IGM) [20]. VDN has been one of the first methods to extend Q-Learning to MARL using CTDE [21]. It implements a not-parameterized CF which computes the $Q_{tot}$ of the joint action as the sum of the individual agents' action-values. Despite respecting IGM, this CF is too simple to model effectively the agents' contribution to $Q_{tot}$ [18].

QMix [18] demonstrated that in order to satisfy IGM, it is sufficient that the CF is monotonic with regard to the individual action-values. As a result, the VDN's sum-function is substituted with a multi-layer perceptron (MLP). This *mixer network* can learn sophisticated non-linear projections of several action-values over $Q_{tot}$. Its weights are generated by a set of hypernetworks conditioned by the state $s$ and are forced to be positive by an absolute activation function. Our proposed method is a refined version of QMix. In particular, TransfQMix also learns a monotonic CF conditioned by $s$ that serves to produce $Q_{tot}$ from the individual action-values. Nonetheless, TransfQMix is a much more sophisticated method for the use of transformers.

Previous methods have attempted to improve QMix. OWQMix and CWQMix [17] used a weighting mechanism for learning non-monotonic CFs, giving more importance to better joint actions. QTran [20] learned a factorization of $Q_{tot}$ that was also free of monotonicity, but it did this via several MLPs. QPlex [26] proposed a dueling structure to learn non-monotonic CFs while adhering to IGM principle. Notice that QPlex, like TransfQMix, employed multi-head attention, but only for a subset of their centralized dueling network. All of these approaches involved RNN agents and large concatenated observation vectors. Despite showing significant advantages in simple theoretical frameworks, it is still debated whether relaxing the monotonicity constraint benefits modeling complex problems [26]. Our refinement of QMix focuses on the representation of cooperative games and the networks architecture rather than monotonicity.

Transformers were successfully deployed in single agent RL [16], but required architecture modifications. Such adjustments are unnecessary for multi-agent problems, since they can be represented more naturally as graph problems. DeepMind's generalist agent (Gato) [19] is a standard transformer that can solve a variety of RL tasks, but it has not been tested in multi-agent settings. Furthermore, Gato is not trained using RL, but rather through a supervised approach. In a method known as universal policy decomposition transformer (UPDET) [5], transformers were applied to a subset of the SC2 tasks. UPDET adopted the QMix framework but replaced RNN agents with transformers, and used a decoupling policy system in which the q-values of entity-based actions (particularly, the q-value of attacking a specific enemy in SC2) were generated by the transformer embedding of that entity. The model performed well in the SC2 subset, but it was not stated how it could be applied to other MARL problems. Moreover, the authors demonstrate that, in the absence of the decoupling approach, QMix performed better when utilizing RNN rather than transformers. Because policy decoupling is not applicable in many scenarios, UPDET appears to be effective only for very specific problems.

Our method formalizes a generic framework that shows clear benefits of using transformers also when policy decoupling is not applicable. TransfQMix employees a transformer also in the central mixer, whereas UPDET deploys the same MLPs of QMix. This makes TransfQMix a totally transferable method. In contrast, UPDET is only partially transferable, because the mixer network must be retrained every time the agents are applied to a new task. TransfQMix uses a recurrent graph approach similar to the one introduced by UPDET. However, TransfQMix makes a better use of the hidden-state by sampling the non-decoupled actions directly from it. Moreover, TransfQMix employs this recurrent mechanism as well in the mixer network. To our knowledge, this is the first method that includes a temporal conditioning in a CF.

Zero-shot transfer, transfer learning, and curriculum learning were explored in MARL by [1] using an entity-based graph method similar to ours. That technique, however, was limited to communication problems, whereas TransfQMix aims to be a general MARL method.

## 3 BACKGROUND

Cooperative multi-agent tasks are formalized as *decentralised partially observable Markov decision process* (Dec-POMDP) [13]. A tuple $G = \langle S, U, P, r, Z, O, H, n, \gamma \rangle$ describes the agents $a \in A \equiv \{1, \ldots, n\}$ which at every time step choose an action $u^a \in U$ from their hidden state $h^a \in H$, forming a joint action $\mathbf{u} \in \mathbf{U} \equiv U^n$. This causes a transition on the environment according to the state transition function $P(s' \mid s, \mathbf{u}) : S \times \mathbf{U} \times S \to [0, 1]$, where $s \in S$ is the true state of the environment. All agents share the same reward function $r(s, \mathbf{u}) : S \times \mathbf{U} \to \mathbb{R}$ and $\gamma \in [0, 1)$ is a discount factor. The agents have access only to partial observations of the environment, $z \in Z$ according to the observation function $O(s, a) : S \times A \to Z$. Each agent has an action-observation history $\tau^a \in T \equiv (Z \times U)^*$, on which it conditions a stochastic policy $\pi^a(u^a \mid \tau^a) : T \times$ $U \to [0, 1]$. The joint policy $\pi$ has a joint action-value function: $Q^\pi(s_t, \mathbf{u}_t) = \mathbb{E}_{s_{t+1:\infty}, \mathbf{u}_{t+1:\infty}}[R_t \mid s_t, \mathbf{u}_t]$, where $R_t = \sum_{i=0}^\infty \gamma^i r_{t+i}$ is the discounted return.

In order to find the optimal joint action-value function $Q^*(s, \mathbf{u}) = r(s, \mathbf{u}) + \gamma \mathbb{E}_{s'}[\max_{\mathbf{u}'} Q^*(s', \mathbf{u}')]$, we use Q-Learning [28] with a deep neural network parameterized by $\theta$ [23] to minimize the expected TD error [26]:

$$\mathcal{L}(\theta) = \mathbb{E}_{(\tau, \mathbf{u}, r, \tau') \in D}\left[\left(r + \gamma V(\tau'; \theta^-) - Q(\tau, \mathbf{u}; \theta)\right)^2\right] \quad (1)$$

where $V(\tau'; \theta^-) = \max_{\mathbf{u}'} Q(\tau', \mathbf{u}'; \theta^-)$ is the one-step expected future return of the TD target and $\theta^-$ are the parameters of the target network, which will be periodically updated with $\theta$. We use a buffer $D$ to store the transition tuple $(\tau, \mathbf{u}, r, \tau')$, where $r$ is the reward for taking action $\mathbf{u}$ at joint action-observation history $\tau$ with a transition to $\tau'$.

We adopt a monotonic CTDE learning paradigm [4, 7, 18, 21]. Execution is decentralized, meaning that each agent's learnt policy is conditioned only on its own action-observation history $\tau^a$. During training, a central mixer network has access to the global state $s$ of the environment and the hidden states of the agents $H$ for projecting the individual action-values over the $Q_{tot}$ of the joint action, which is used in equation (1) to train the model end to end. The monotonic constraint imposed to the CF is the same formalized by QMix:

$$\frac{\partial Q_{tot}}{\partial Q_a} \geq 0, \forall a \in A \quad (2)$$

which ensures that the IGM principle is respected.

The neural networks in our method are transformers [24], which make a large use of the attention mechanism [2]. Specifically, we use transformers to manipulate our graphs via *multi-head self-attention* (MHSA) [10, 15, 24]. Given an embedded graph matrix $\mathbf{X}^{n \times h}$ of $n$ vertices represented with $h$-dimensional vectors, a transformer computes a set of queries $\mathbf{Q} = \mathbf{X}\mathbf{W}^Q$, keys $\mathbf{K} = \mathbf{X}\mathbf{W}^K$, and values $\mathbf{V} = \mathbf{X}\mathbf{W}^V$, where $\mathbf{W}^Q, \mathbf{W}^K, \mathbf{W}^V$ are three different parameterized matrices with dimensions $h \times k$. The self-attention is then computed as:

$$\text{Self-Attention}(\mathbf{X}) = \text{Attention}(\mathbf{Q}, \mathbf{K}, \mathbf{V}) = \text{softmax}\left(\frac{\mathbf{Q}\mathbf{K}^\top}{\sqrt{n}}\right)\mathbf{V} \quad (3)$$

A transformer uses $m$ attention modules in parallel, and then concatenates all the outputs and projects them back to $h$-dimensional vectors using a final $\mathbf{W}^O$ feed-forward layer:

$$\text{MultiHeadSelfAttn}(\mathbf{X}) = \text{Concat}(\text{head}_1, \cdots, \text{head}_m)\mathbf{W}^O$$
$$\text{where } \text{head}_i = \text{Attention}\left(\mathbf{X}\mathbf{W}_i^Q, \mathbf{X}\mathbf{W}_i^K, \mathbf{X}\mathbf{W}_i^V\right). \quad (4)$$

## 4 METHOD

### 4.1 Graph Observations and State

Our method rethinks how cooperative problems are presented to neural networks. For the sake of simplicity, here we assume that an agent observes $k$ entities at each time step $t$, where $k$ is the total number of entities in the environment. In our approach, a set of $z$ features defines each entity. Because of the environment's partial observability, the features can take different values for each agent.

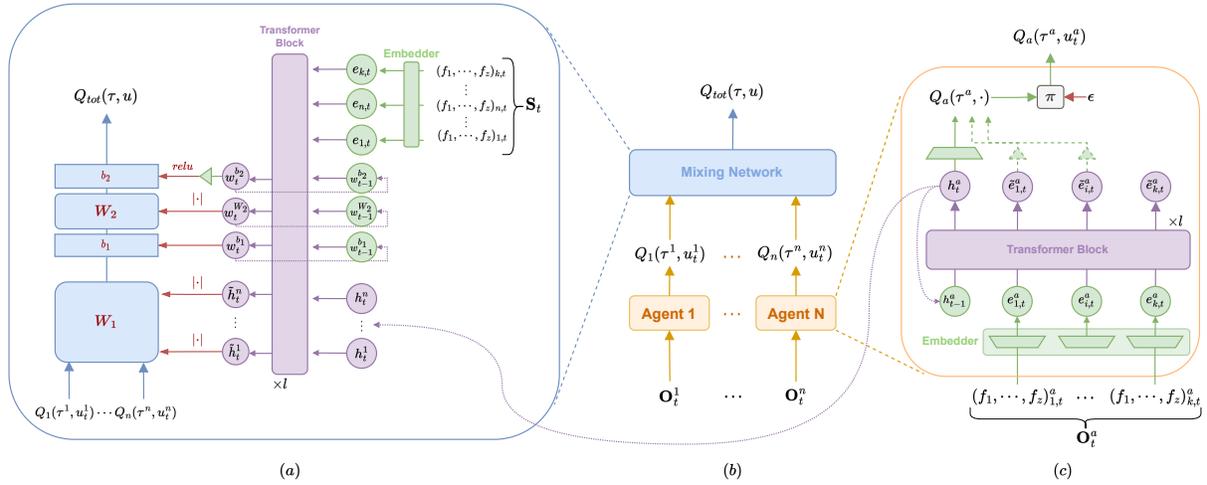

Figure 2: (a) Transformer Mixer. (b) Overall TransfQMix architecture. (c) Transformer Agent. The purple dotted lines represent the recurrent connections. The green components are simple feed-forward layers (embedders and scalar projectors), and the green circles are the embedded vertices. The purple circles are transformed vertices. The dotted green components represent the action decoupling mechanism.

Therefore,

$$ent_{i,t}^a = [f_1, \cdots, f_z]_{i,t}^a \quad (5)$$

defines the entity $i$ as it is observed by the agent $a$ at the time step $t$. We replace the traditional observation vectors with observation matrices with dimensions $k \times z$ which includes all the $k$ entities observed by an agent $a$ at $t$:

$$\mathbf{O}_t^a = \begin{bmatrix} ent_1 \\ \vdots \\ ent_k \end{bmatrix}_t^a = \begin{bmatrix} f_{1,1} & \cdots & f_{1,z} \\ \vdots & \ddots & \vdots \\ f_{k,1} & \cdots & f_{k,z} \end{bmatrix}_t^a \quad (6)$$

This structure allows the agents to process the features of the same type using the same weights of a parameterized matrix **Emb** with shape $z \times h$, where $h$ is an embedding dimension. The resulting matrix $\mathbf{E}_t^a = \mathbf{O}_t^a \mathbf{Emb}^a$ is formed by vertices embeddings $[e_1, \cdots, e_k]_t^{a\top}$ that will be further processed by transformers. Notice that $\mathbf{Emb}^a$ is independent from $k$. Conversely, the encoding feed-forward layer used by RNN agents has approximately $k \times z \times h$ parameters. Our approach is therefore more scalable and transferable in respect to the number of entities.

The observation vectors in the cooperative environments we studied [11, 20, 25] already contained an implicit matrix structure or required very little modifications to adopt it. Features like (relative) map location, velocity, remaining life points, and so on, which are frequently defined for all entities and then concatenated in the same vector, can be easily rethought as vertex features of our observation matrix. On the other hand, features such as one-hot-encoding of agent's last action or one-hot-encoding of agent's id necessitate extra work. Moreover, since in our method the features of the same types are processed by the same weights, we lose the *positional* information implicitly present in the concatenated vectors. A traditional encoder, indeed, can learn that the features in some specific vector locations are relevant to some specific entity and hence treat them differently from the others.

In our preliminary research, we found that we can compensate for these drawbacks by using two additional binary features. The first, IS_SELF, informs if the described entity is the agent to which the observation matrix belongs:

$$f_{i,\text{IS\_SELF}}^a = \begin{cases} 1, & \text{if } i = a \\ 0, & \text{otherwise.} \end{cases} \quad (7)$$

This feature will be 1 for $ent_{a,t}^a$ and 0 for all the other entities. IS_SELF can be thought as a re-adaptation of the one-hot-encoding of the agent's id, which is commonly employed by state-of-the-art models [18, 26, 30]. The second feature tells us if the entity described is a cooperative agent or not:

$$f_{i,\text{IS\_AGENT}}^a = \begin{cases} 1, & \text{if } i \in A \\ 0, & \text{otherwise} \end{cases} \quad (8)$$

allowing the vertex features of teammates to be treated differently than others. Even though state-of-the-art methods do not always include this feature, we argue that we are not using additional data because this information is otherwise implicitly encoded in vector positions.

We apply the same reformulation of the agents' observations to the global state. Usually, the state is defined as a vector of "real" features relative to the entities (i.e., not partially observed by an agent) and/or the concatenation of all agents' observations. In our approach, we define a state matrix $\mathbf{S}_t$ of dimensions $k \times z$:

$$\mathbf{S}_t = \begin{bmatrix} ent_1 \\ \vdots \\ ent_k \end{bmatrix}_t = \begin{bmatrix} f_{1,1} & \cdots & f_{1,z} \\ \vdots & \ddots & \vdots \\ f_{k,1} & \cdots & f_{k,z} \end{bmatrix}_t \quad (9)$$

which defines the vertex features for all the entities from a global point of view. For simplicity, in the notation we assume that we are using the same $z$ features in both **S** and **O**. We could use different ones, though. For instance, adding IS_SELF to **S** does not make sense since the features are not defined in respect to any agent, and indeed in our experiments we do exclude IS_SELF from S. In the environments that we took into account, the state vectors shown a structure easily reshapable as in equation (9). As for O, we can process the same feature types in parallel with a parameterized matrix $\mathbf{Emb}^s$ to obtain embedded vertices that can be further processed by a transformer, i.e. $\mathbf{E}_t = [e_1, \cdots e_k]_t^\top = \mathbf{S}_t \mathbf{Emb}^s$.

### 4.2 Transformer Agent

Our transformer agent takes as input the embedded vertices $\mathbf{E}_t^a = [e_1, \cdots, e_k]_t^{a\top}$ plus a hidden vector $h_{t-1}^a$, which has the same size of any vector $e_i^a$ and it is fullfilled with 0s at the beginning of an episode. The final input matrix is $\mathbf{X}_t^a = [h_{t-1}^a, e_{1,t}^a, \cdots, e_{k,t}^a]^\top$. The output of $l$ transformer blocks: $\tilde{\mathbf{X}}_t^a = \text{MultiHeadSelfAttn}(\mathbf{X}_t^a)$ is a refined graph in which all the vertices were altered based on the attention given to the others. In particular, $h_t^a = \tilde{h}_{t-1}^a$ can be considered as a transformation of the agent's hidden state according to the attention given to the new state of the entities. Similarly to the approach used in natural language processing, where the transformation of the first token ( [CLS] in Bert [3]) is considered to encode an entire sentence, we consider $h_t^a$ to encode the general coordination reasoning of an agent. We therefore sample the agent's actions-values from $h_t^a$ using a feed-forward layer $\mathbf{W}^u$ with dimensions $h \times u$, where $u$ is the number of actions: $Q_a(\tau^a, \cdot) = h_t^a \mathbf{W}^u$. Finally, we pass $h_t^a$ to the next time step so that the agent can update its coordination reasoning recurrently. When some agent's actions are directly related to some of the observed entities (e.g., "attack the enemy $i$" in StarCraft II), our transformer agents use a decoupling mechanism similar to the one introduced in [5]. In particular, the action-values of the entity-related actions are derived from their respective entity embeddings. An additional feed-forward matrix $\mathbf{W}^{\hat{u}}$ of dimension $h \times 1$ is used in this case. For example, the q-value of attacking the enemy $i$ is sampled as $\tilde{e}_{i,t}^a \mathbf{W}^{\hat{u}}$. The q-values of the non-entity-related and the entity-related actions are then concatenated together to obtain $Q_a(\tau^a, \cdot)$.

### 4.3 Transformer Mixer

Exactly as QMix, TransfQMix uses a MLP in order to project $Q_A$ (the q-values of the actions sampled by the individual agents) over $Q_{tot}$ (the q-value of the joint sampled action). Formally:

$$Q_{tot} = (Q_A^{(1 \times n)} \mathbf{W}_1^{(n \times h)} + \mathbf{b}_1^{(1 \times h)}) \mathbf{W}_2^{(h \times 1)} + \mathbf{b}_2^{(1 \times 1)} \quad (10)$$

where $\mathbf{W}_1, \mathbf{b}_1$ and $\mathbf{W}_2, \mathbf{b}_2$ are the weights and biases of the hidden and output layer, respectively. We explicitly state inside brackets the dimensions of equation 10 to show that only three values are relevant: $n$, the number of agents; $h$, a hidden dimension; and 1, which accounts for $Q_{tot}$ being a scalar. This shows that in order to arrange the MLP mixer we need $n + 2$ vectors of size $h$ plus a scalar.

QMix generates the vectors using 4 MLP hypernetworks. We propose to use the outputs of a transformer to generate the weights of the mixer's MLP. The input graph of our transformer mixer is:

$$\mathbf{X}_t = \left[h_t^1, \cdots, h_t^n, w_{t-1}^{\mathbf{b}_1}, w_{t-1}^{\mathbf{W}_2}, w_{t-1}^{\mathbf{b}_2}, e_{1,t}, \cdots, e_{k,t}\right]^\top \quad (11)$$

where $h_t^1, \cdots, h_t^n$ are the $n$ hidden states of the agents, $w_{t-1}^{\mathbf{b}_1}, w_{t-1}^{\mathbf{W}_2}, w_{t-1}^{\mathbf{b}_2}$ are three recurrent vectors fulfilled with 0s at the beginning of an episode, and $e_{1,t}, \cdots, e_{k,t}$ is the embedded state, i.e., $\mathbf{E}_t = \mathbf{S}_t \mathbf{Emb}^s$. The output consist in a matrix $\tilde{\mathbf{X}}_t = \text{MultiHeadSelfAttn}(\mathbf{X}_t)$ that contains the same vertices of $\mathbf{X}_t$ transformed by the multi-head self-attention mechanism. In particular, $\tilde{h}_t^1, \cdots, \tilde{h}_t^n$ are the coordination reasonings of agents enhanced by global information to which the agents had no access, namely the hidden state of the other agents and the true state of the environment. These $n$ refined vectors are used to build $\mathbf{W}_1$. $Q_A \mathbf{W}_1$ is therefore a re-projection of the individual q-values $Q_A$ over a transformation of the agents' hidden states. Notice that the individual q-values were generated (or conditioned) exactly from $h_t^1, \cdots, h_t^n$ by the agents. This means that the primary goal of the transformer mixer is to combine and refine the independent agents' reasoning so that they represent the team coordination.

The transformed embeddings of the recurrent vectors, $w_t^{\mathbf{b}_1} = \tilde{w}_{t-1}^{\mathbf{b}_1}, w_t^{\mathbf{W}_2} = \tilde{w}_{t-1}^{\mathbf{W}_2}, w_t^{\mathbf{b}_2} = \tilde{w}_{t-1}^{\mathbf{b}_2}$ are used to generate $\mathbf{b}_1, \mathbf{W}_2, \mathbf{b}_2$, respectively. Since $\mathbf{b}_2$ is a scalar, an additional parameterized matrix with dimensions $h \times 1$ is applied on $w_t^{\mathbf{b}_2}$. We use a recurrent mechanism for two reasons: (i) to ensure that the transformer mixer is totally independent of the number of entities in the environment; and (ii) to incorporate a temporal dependence on the centralized training, in accordance with the MDP formulation of the problem. We argue that $Q_{tot}$ is heavily dependent on prior states and that this reliance should be encoded explicitly on the mixer network. This recurrent process allows the mixer to provide more consistent targets across time steps, resulting in more stable training.

We employ the same strategy described by QMix to adhere to the monotonicity constraint. Namely, we apply an absolute activation function to the weights $\mathbf{W}_1$ and $\mathbf{W}_2$ and $relu$ to $\mathbf{b}_2$.

## 5 EXPERIMENTAL SETUP

### 5.1 Spread

In the Spread environment [11], the goal of $n$ agents is to move as close as possible to the random positions occupied by $n$ landmarks while avoiding collisions with each other. The agents can move in four directions or stay still. The optimal policy would have one agent occupying one landmark, resulting in a perfect space distribution. Since each agent must anticipate which target the other agents will occupy and proceed to the remaining one, this calls for robust coordination reasoning.

The global reward is the negative minimum distances from each landmark to any agent. An additional term is added to punish collisions among agents. It must be noticed that the original reward function implemented by [11] was affected by a redundant factor, i.e. it is multiplied by $2n$. Later on, PettingZoo [22] eliminated this redundancy, which is the reward function we used here.

The Spread's observation space for the agent $a$ consists of a vector containing the velocity and absolute position of itself together with the relative positions of all the other agents and landmarks. In order to convert it into an observation matrix, we only maintain the relative positions, which are the features defined for all the entities observed by $a$. Every observed entity is therefore defined by $ent_{i,t}^a = [pos_x, pos_y, \text{IS\_SELF}, \text{IS\_AGENT}]_{i,t}^a$ where $pos_x$ and $pos_y$ are the relative positions of the entity $i$ in respect to $a$ in the horizontal and vertical axes.

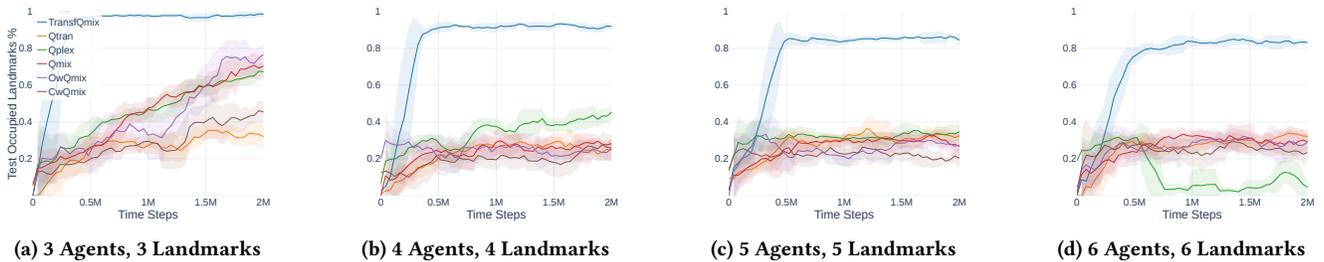

Figure 3: Comparative results in the Spread environment.

The Spread's state space consist of the concatenation of all the agents observations. Also in this case we keep the features that are defined for all the entities, which are the absolute positions and the velocities. In the final state matrix the entities are defined by $ent_{i,t} = [\hat{pos}_x, \hat{pos}_y, v_x, v_y, \text{IS\_AGENT}]_{i,t}$ where $\hat{pos}_x$ and $\hat{pos}_y$ are the absolute position of the entity $i$, and $v_x$ and $v_y$ its velocity (which is 0 in the case of the landmarks).

The standard reported metric for Spread is the global reward. This metric, however, is not informative because it is a value that is challenging to interpret and does not stay in the same range when $n$ changes. As a result, we present a new metric: the percentage of landmarks occupied at the conclusion of an episode (POL). To compute the POL we count the number of landmarks with an agent closer than a predetermined threshold and we divide it for the total number of landmarks. The POL is a more informative metric because it assesses the proper distribution of the agents. Additionally, it maintains the same range (0, 1) when $n$ is changed. We found that when the distance threshold is set to 0.3, the POL has a correlation of 0.95 with the reward function, meaning that the data we are presenting is still comparable with previous studies.

### 5.2 StarCraft II

This environment uses the StarCraft II Learning Environment [25], which makes available a range of micromanagement tasks based on the well-known real-time strategy game StarCraft II[1]. Each task consists of a unique combat scenario in which a group of agents, each managing a single unit, engage an army under the command of the StarCraft game's central AI. In order to win a game, agents must develop coordinated action sequences that will allow them to concentrate their attention on certain enemy units. We report the results in SC2 for the 8 tasks that are considered the most difficult in the literature [26, 30]: 5m_vs_6m, 8m_vs_9m, 27m_vs_30m, 5s10z, 3s5z_vs_3s6z, 6h_vs_8z, MMM2, and corridor.

The SC2's observation vector for the agent $a$ consists in a concatenation of features defined for the allies and the enemies that are inside the *sight range* of the agent. These features include the relative position of the entity in respect to $a$, the distance, the health, the state of the shield, and a one-hot-encoding of the type of the entity (which can be a marine, a marauder, a stalker, etc.). This structure already defines an observation matrix which requires only the addition of the IS_SELF and IS_AGENT features to be used by TransfQMix. However, TransfQMix can not use some additional features that are present in the original SC2's observation vector, which include a one-hot-encoding of the available and previous actions and a representation of the map's limits.

---
[1]StarCraft II is a trademark of Blizzard Entertainment[TM]

Our transformer mixer can be fed directly with the original state vector of SC2, which is also a concatenation of features defined for all $k$ entities. These features are the same of the observation vector but defined from a global viewpoint, i.e., the position relative to the center of the map. On the other hand, an additional feature consisting of the actions taken by all the agents is not used by TransfQMix since it is not compatible with the graph state approach.

The decoupling technique described in Section 4.2 is employed for TransfQMix and UPDET, i.e., the q-value of attacking the enemy $i$ is determined from the transformer embedding of $i$. When appropriate, the same process is used for actions that include healing another agent.

### 5.3 Algorithms

Our codebase is built on top of pymarl [4, 18] and it is available at: https://github.com/mttga/pymarl_transformers. It contains TransfQMix and our wrappers for Spread and SC2, plus the original implementations of the algorithms to which our method is compared to: QMix, QTran, QPlex, OW-QMix, CW-QMix and UPDET. For all the compared methods, we used the same hyper-parameters reported in the original implementations. We kept the parameters of each method constant in all the experiments we performed. Notice that in all our experiments we used the *parameters sharing* technique, i.e., all the agents shared the same weights. This was demonstrated to be very beneficial in several studies [14, 18, 30].

We fine-tuned TransfQMix in the SC2's 5m_vs_6m task and we used the same parameters for all the other settings (included the Spread tasks). In particular, we used 32 as the hidden embedding dimension, 4 attention heads, and 2 transformer blocks for both the transformer agents and the mixer, resulting in a total of $\sim 50k$ parameters for both networks. The learning configuration for all the transformers architectures (included UPDET) used the *Adam* optimizer with a learning rate of 0.001 and a $\lambda$ of 0.6 for computing the twin delayed (td) targets. This setup is different from the one used by the state-of-the-art RNN-based models (*RMSProp* optimizer, 0.0005 learning rate, and 0 for td's $\lambda$). However, we found that the optimal learning configuration of TransfQMix did not work with the other models, i.e., they performed better with their original learning setup. Some parameters were shared by all the methods, such as the buffer size (5000 episodes), the batch size (32 episodes), the interval for updating the target network (200 episodes), and the anneal time for the epsilon decay (100$k$ time steps).

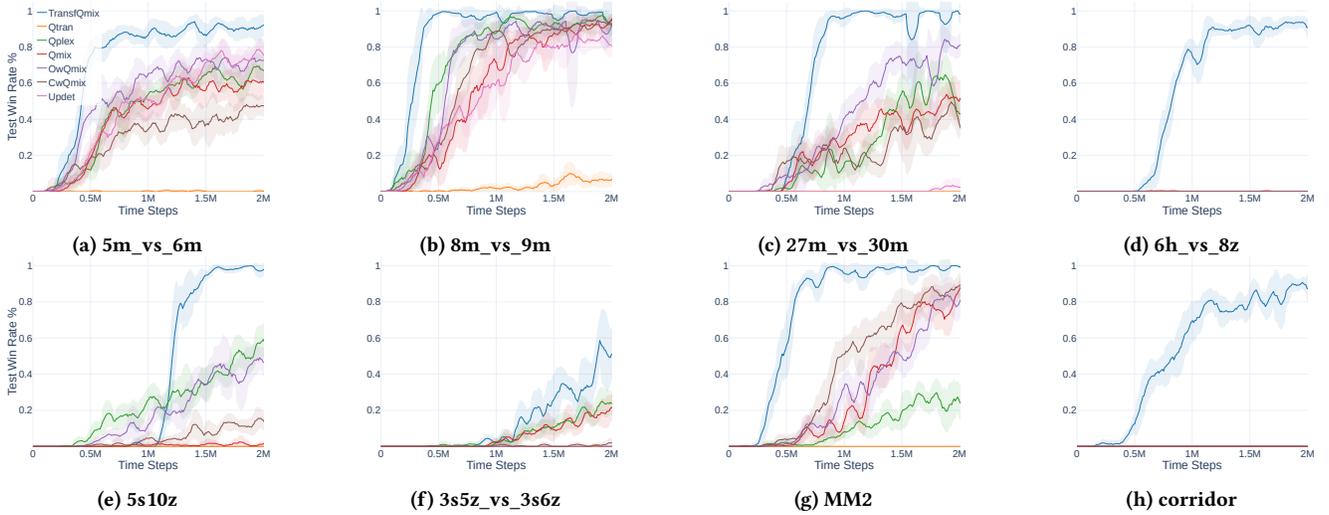

Figure 4: Comparative results in the SC2 environment.

## 6 RESULTS AND DISCUSSION
### 6.1 Main Results

The performances of MARL methods in Spread are usually reported using $n = 3$. We increased $n$ up to 6 in order to analyze the scalability of the methods. Figure 3 shows how the POL improved when the considered methods were trained on the various scenarios. The POL was computed every $40k$ time steps by running 30 independent episodes with each agent performing greedy decentralised action selection. In the standard task involving 3 agents, state-of-the-art methods learned a good policy which covers on average the ∼80% of the landmarks, with the exception of QTran and CW-QMix (POL of ∼50%). However, they did not perform significantly better than QMix. The sole state-of-the-art method that could defeat QMix in the tasks involving 4 or 5 agents was QPlex (POL of 50%), which demonstrated to be very unstable in $n = 6$. Conversely, TransfQMix significantly outperformed QMix and the other methods in every scenario, reaching a steady POL of almost 90% in just $500k$ time steps. Notice that in Spread the optimal policy was the same for every $n$ (i.e., each agent occupying a landmark). State-of-the-art methods could learn this strategy only when the team size was small. On the other hand, TransfQMix demonstrated a better agent-team size invariance by obtaining similar results in every scenario.

Figure 4 shows the results of all the methods in the hardest tasks of SC2. The reported metric is the average percentage of won games performing greedy action sampling every 100 episodes during training. The results for UPDET are reported only for tasks that include marines, since the original implementation of this method does not support other scenarios. It is noteworthy that UPDET did not perform better than RNN-based models and failed in the 27m_vs_30m task, indicating that using a transformer agent with policy decoupling does not necessarily provide a clear advantage. Conversely, our more sophisticated use of transformers significantly outperformed the other models in every task, and consistently defeats the SC2's central AI even in scenarios where previous methods could not win any game. TransfQMix also demonstrated its effectiveness in environments with a large number of entities, such as the corridor map (which stays for 6 Zealots versus 24 Zerglings, for a total of 30 entities) and the 27m_vs_30m (57 entities). While other approaches require their parameters to be increased according to the number of entities, TransfQMix's networks are (nearly) the same size in all the tasks. This suggests that TransfQMix's architecture may be regarded as sufficiently generic to address various problems without requiring structural changes.

Table 1: Results of zero-shot transfer in Spread.

|  | POL Scenario | | | |
| Model | 3v3 | 4v4 | 5v5 | 6v6 |
| --- | --- | --- | --- | --- |
| TransfQMix (3v3) | **0.98** | 0.88 | 0.8 | 0.75 |
| TransfQMix (4v4) | 0.96 | **0.93** | **0.9** | 0.86 |
| TransfQMix (5v5) | 0.88 | 0.85 | 0.82 | 0.82 |
| TransfQMix (6v6) | 0.91 | 0.88 | 0.85 | 0.84 |
| TransfQMix (CL) | 0.88 | 0.88 | 0.87 | **0.87** |
| State-of-the-art | 0.76 | 0.45 | 0.36 | 0.33 |

### 6.2 Transfer Learning

We tested the zero-shot capabilities of TransfQMix by applying the networks trained in a particular Spread task to the others. Table 1 shows the POL averaged across 1000 episodes achieved by TransfQMix trained with $n$ agents in scenarios with different $n$. As a benchmark, the best POLs obtained by state-of-the-art models trained in each specific task are reported. We also include the performances of TransfQMix trained with a curriculum learning (CL) approach, which consists of making the agents cooperate in progressively larger teams. In particular, we trained the agents in teams of 3, 4, 5 and 6 for $500k$ time steps each.

In general, every network showed excellent zero-shot capabilities but worse performances for larger teams, except for the agents trained with CL, which performed similarly in all the scenarios.

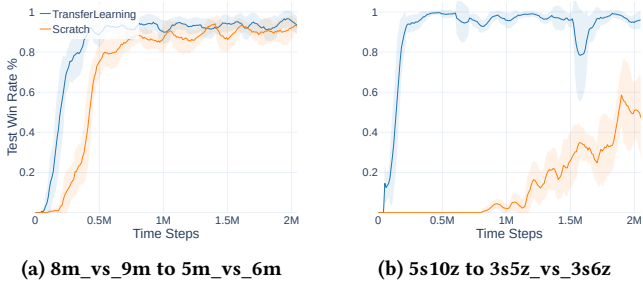

(a) 8m_vs_9m to 5m_vs_6m

(b) 5s10z to 3s5z_vs_3s6z

Figure 5: Transfer learning vs training from scratch in 2 SC2 tasks.

In this sense, CL seems a promising approach for obtaining general coordination policies with TransfQMix. Surprisingly, the best transferable policy was learned in the 4v4 task. This could be because the scenario is complex enough to necessitate the learning of strong coordination policies, but not so complicated as to produce instabilities or slow down the learning process. Finally, it is remarkable that all the TransfQMix's zero-shot transfers outperformed state-of-the-art methods trained in the various scenarios.

The only constraint to using TransfQMix in different contexts is that the vertex feature space must be the same. This is not always guaranteed in the SC2 environment, because the unit type's one-hot encoding feature is dependent on the total number of unit types of the scenario. Nonetheless, we can utilize the same networks in maps with the same entity type. Figure 5 shows the results obtained in the 5m_vs_6m and 3s5z_vs_3s6z tasks by fine-tuning the agents trained in the 8m_vs_9m and 5s10z scenarios, respectively.

In both cases, we are transferring coordination strategies learnt in simpler settings to more difficult tasks, implying that we are conducting a minimal CL. We can see how fine-tuning helped TransfQMix develop a significantly better policy (Figure 5b) or converge faster (Figure 5a) than when it was trained from scratch. The initial peak in the figures corresponds to the zero-shot performance, and it is followed by a falling phase in which the weights were rapidly adjusted for the new task.

In conclusion, the results demonstrate TransfQMix's promising capacity to transfer knowledge between scenarios, as well as how transfer and curriculum learning could aid in the resolution of complex MARL tasks.

### 6.3 Ablation

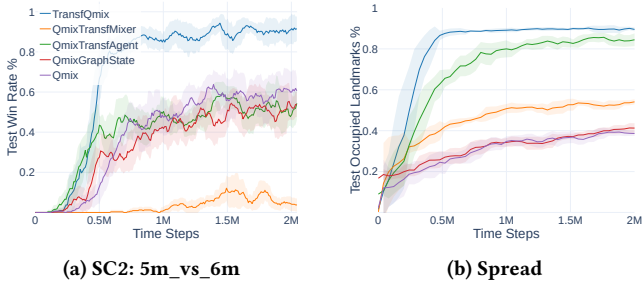

(a) SC2: 5m_vs_6m

(b) Spread

Figure 6: Ablation Study.

It might be claimed that the results obtained by TransfQMix in Spread are not comparable with the previous methods because we employ a state observation matrix that differs from the original state vector. To test this argument, we ran QMix with the flattened version of our state matrix in Spread. The averaged test POL across all Spread tasks is shown in Figure 6b. We can see that QMix's performance with a graph state (QMixGraphState) was not considerably different from QMix's performance with the original state vector. The same figure shows that replacing the QMix mixer's hypernetworks with a transformer mixer improved performance (QMixTransformerMixer). This indicates that, in order to benefit from a graph-based state, a graph-based network as a transformer should be used. We also provide the results produced by our transformer agents in conjunction with the traditional QMix hypernetworks. This framework clearly outperformed the original one based on RNNs in terms of coordination, but it was not as stable or performant as TransfQMix.

The identical ablation study was carried out in the SC2 5m_vs_6m task (Figure 6a). In this scenario, the transformer agents and mixers alone were unable to increase the performance of QMix, implying that we need to utilize transformers in both the agent and mixer networks in order to leverage the graph structure of the observations and state.

## 7 CONCLUSION

In this paper we proposed a novel graph-based formalization of MARL problems that depicts coordination problems in a more natural way. We introduced TransfQMix, a method based on transformers that makes use of this structure to enhance the coordination reasoning of the QMix's agents and mixer. TransfQMix demonstrated great learning capabilities by excelling in the most challenging SC2 and Spread tasks without the need for task-specific hyperparameter tuning. In contrast to prior approaches that attempted to enhance QMix, TransfQMix does not focus on the monotonicity constraint or other aspects of the learning process. This shows that in order to improve MARL methods, neural networks architectures and environment representations need to receive greater focus.

The application of TransfQMix in transfer learning, zero-shot transfer, and curricular learning yielded promising results. In future research we aim to explore the method's generalization abilities by including several tasks into a single learning pipeline. For instance, we aim to train the same agents to solve all the SC2 tasks. Additionally, we want to investigate the feasibility of transferring coordination policies between MARL domains. Finally, we want to examine in greater detail the influence of multi-head self-attention on coordination reasoning.

## ACKNOWLEDGMENTS

This project has received funding from the EU's Horizon 2020 research and innovation programme under the Marie Skłodowska-Curie grant agreement No 893089. This work acknowledges the 'Severo Ochoa Centre of Excellence' accreditation (CEX2019-000928-S). We gratefully acknowledge the David and Lucile Packard Foundation.

# TransfQMix: Transformers for Leveraging the Graph Structure of Multi-Agent Reinforcement Learning Problems (Supplementary Material)

Matteo Gallici
KEMLG Research Group, Universitat Politècnica de Catalunya.
Barcelona, Spain
gallici@cs.upc.edu

Mario Martin
KEMLG Research Group, Universitat Politècnica de Catalunya.
Barcelona, Spain
mmartin@cs.upc.edu

Ivan Masmitja
Institut de Ciències del Mar (ICM), CSIC
Barcelona, Spain
masmitja@icm.csic.es

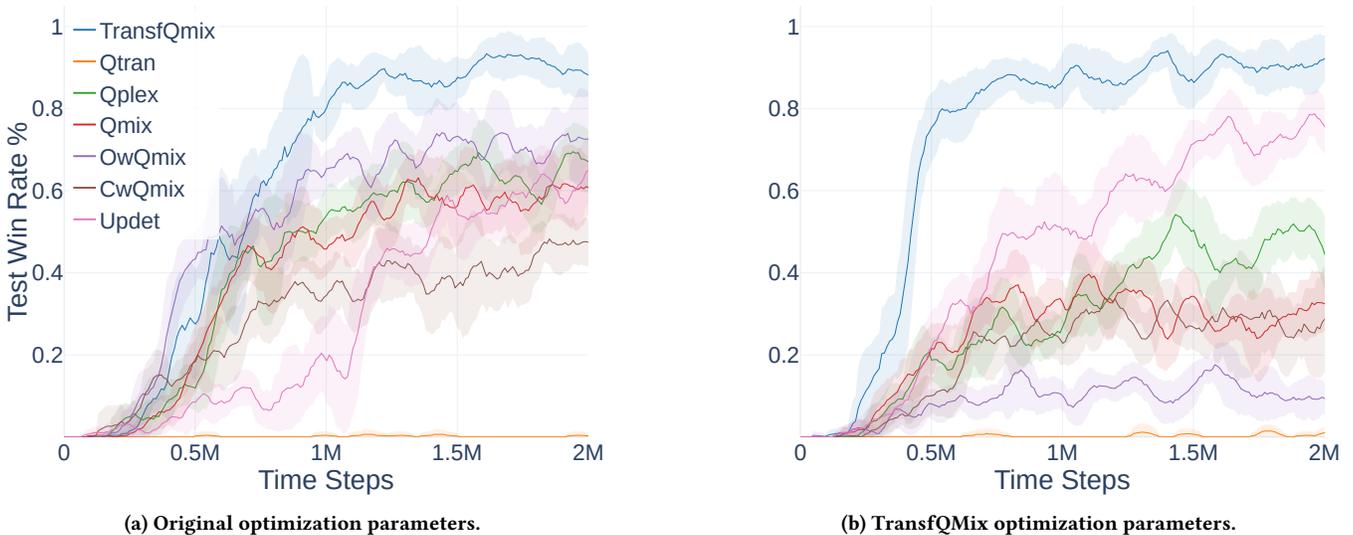

(a) Original optimization parameters.

(b) TransfQMix optimization parameters.

Figure 1: Results obtained in the StarCraft II 5m_vs_6m task using the optimization parameters commonly adopted by state-of-the-art models (*RMSProp* optimizer, 0.0005 learning rate, and 0 for td's $\lambda$) and using the optimal TransfQMix optimization parameters (*Adam* optimizer, 0.001 learning rate, and 0.6 for td's $\lambda$). State-of-the art models do not benefit from the optimization used by TransfQMix. At the same time, TransfQMix outperforms state-of-the-art methods also when it's trained using the state-of-the-art's configuration. Updet is the only method that benefits from using TransfQMix's optimizer configuration, suggesting that these parameters are effective when used to train transformer networks.

Table 1: Parameters of TransfQMix. The parameters relative to the transformer are shared between the transformer agents and the transformer mixer.

| *Parameter* | *Value* | *Description* |
|---|---|---|
| Buffer Size | 5000 | Number of last saved episodes used for training |
| Batch Size | 32 | Batch size used for training |
| Update Interval | 200 | Episode interval for updating the target network |
| Optimizer | Adam | Optimizer |
| Learning Rate | 0.001 | Learning Rate |
| Td-Lambda | 0.6 | Lambda for computing td-targets |
| Emb Dim | 32 | Embedding dimension $h$ |
| Attention Heads | 4 | Self-attention heads of each transformer block |
| Transformer Blocks | 2 | Number of transformer layers |
| Dropout | 0 | Dropout percentage in transformer block |
| Learnable parameters | $\sim 50k$ | Learnable parameters of a single network (mixer or agent) |

Table 2: Comparison between the number of parameters (agent and mixer networks) of TransfQMix and other state of the art models. The number parameters are reported for Spread 3v3, 6v6 and SC2 27m_vs_30m to appreciate their relation with the number of environment's entities. The number of parameters of TransfQMix is invariable in respect to the entities. Conversely, other methods increase their parameters proportionally with the entities, leading to oversized networks in the 27m_vs_30m task of SC2. TransfQMix is on overall a lighter model than other methods (with the exception of QMix in Spread 3v3 and 6v6).

| *Model* | *Agent* | *Mixer* |
|---|---|---|
| TransfQMix | 50k | 50k |
| QMix | 27k | 18k |
| QPlex | 27k | 251k |
| O-CWQMix | 27k | 179k |

(a) Spread 3v3

| *Model* | *Agent* | *Mixer* |
|---|---|---|
| TransfQMix | 50k | 50k |
| QMix | 28k | 56k |
| QPlex | 28k | 597k |
| O-CWQMix | 28k | 301k |

(b) Spread 6v6

| *Model* | *Agent* | *Mixer* |
|---|---|---|
| TransfQMix | 50k | 50k |
| QMix | 49k | 283k |
| QPlex | 49k | 3184k |
| O-CWQMix | 49k | 1021k |

(c) SC2 27m_vs_30m

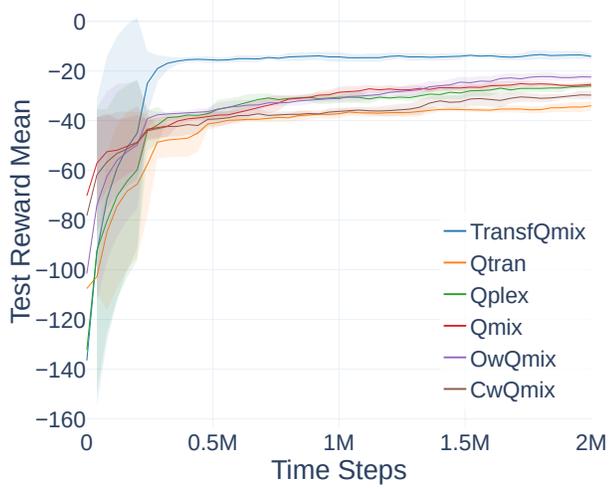
(a) 3 Agents, 3 Landmarks

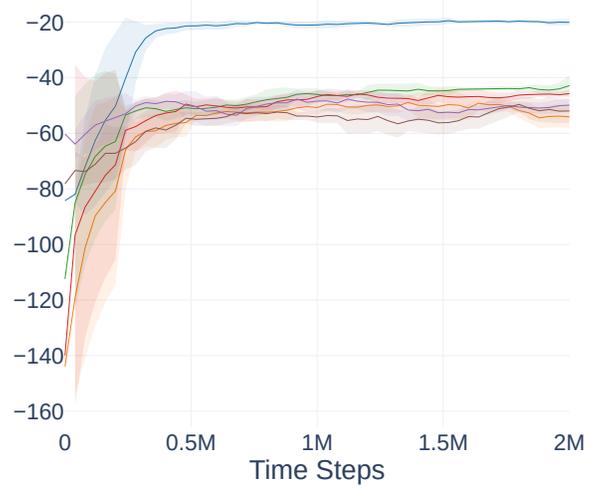
(b) 4 Agents, 4 Landmarks

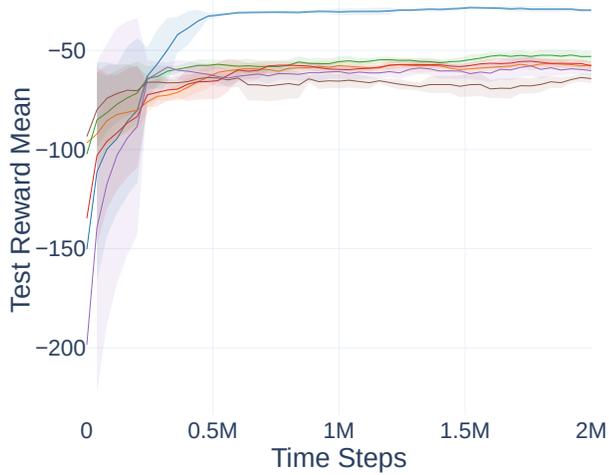
(c) 5 Agents, 5 Landmarks

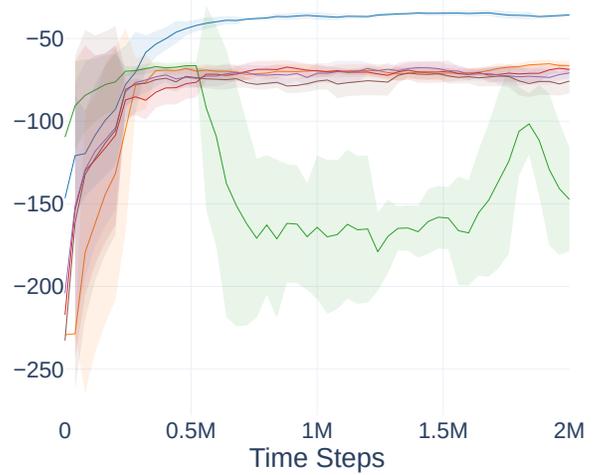
(d) 6 Agents, 6 Landmarks

Figure 2: Average reward in Spread performing greedy action selection during training. The global reward is the negative minimum distances from each landmark to any agent. We used the PettingZoo reward, which is proportional to $1/2n$ in respect to the original one. The results are proportional to the ones based on POL showed in the paper.

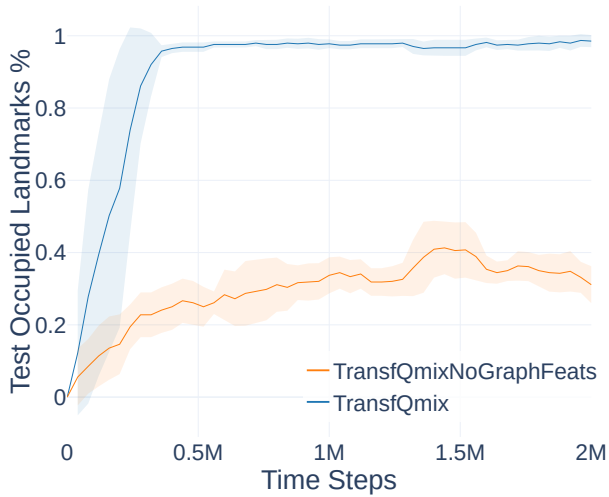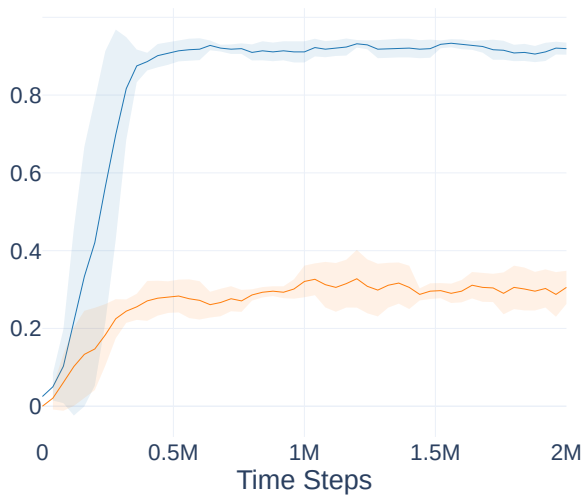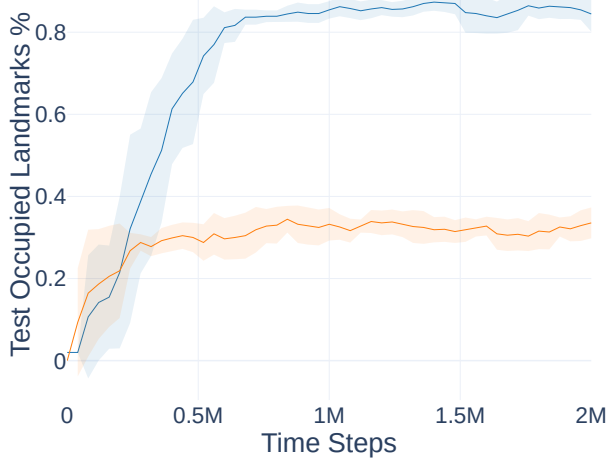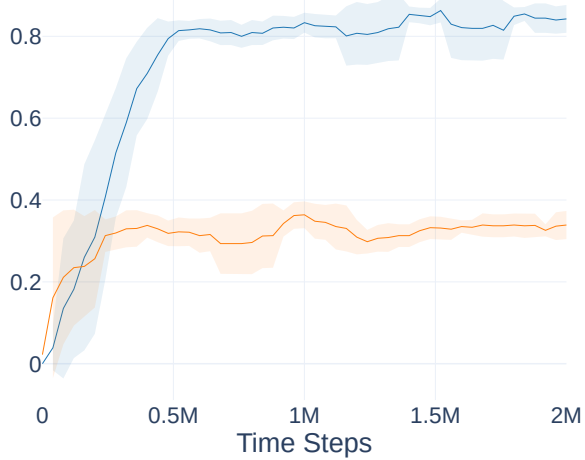

Figure 3: POL in Spread tasks performing greedy action during training of TransfQMix using IS_SELF and IS_AGENT vertex features (TransfQMix) and not using them (TransfQMixNoGraphFeats). Despite how simple they are, these two binary features allow the transformer to infer which of the entity embeddings are relative to the current agent and which of the other ones are relative to team-mates. This seems extremely important in order to generate a coherent coordination graph using self-attention.

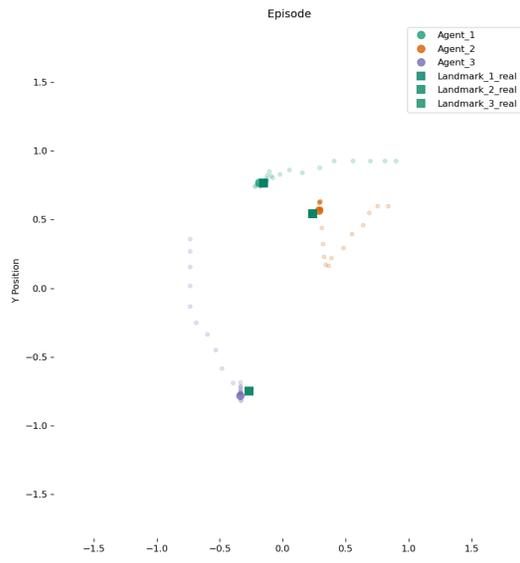
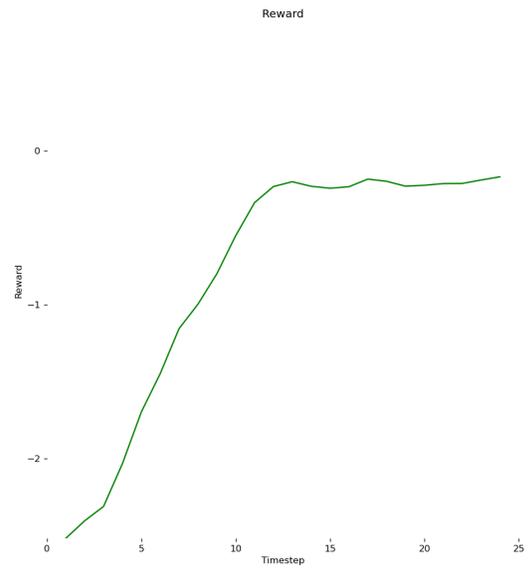

(a) 3 Agents, 3 Landmarks

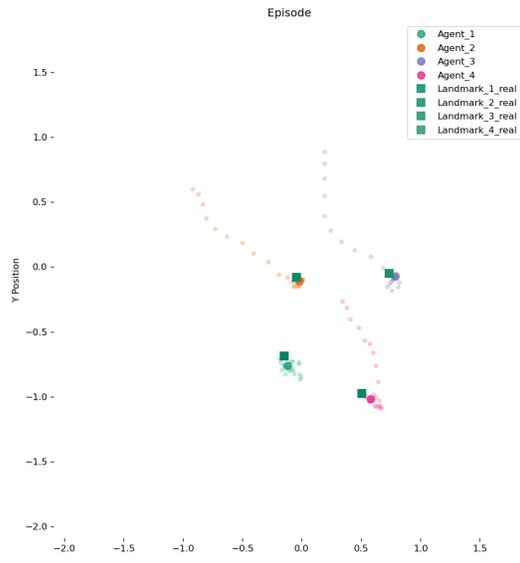
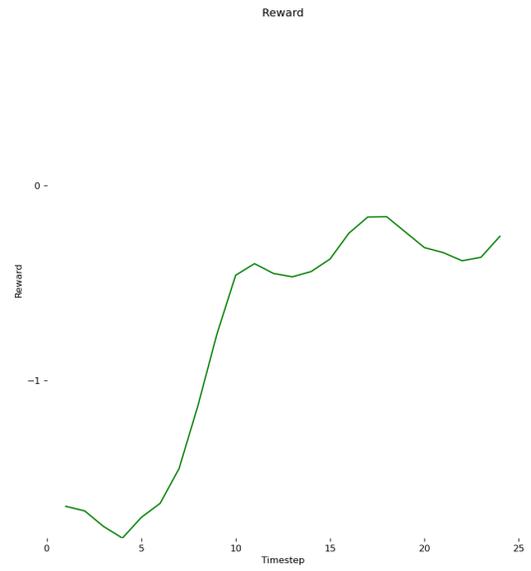

(b) 4 Agents, 4 Landmarks

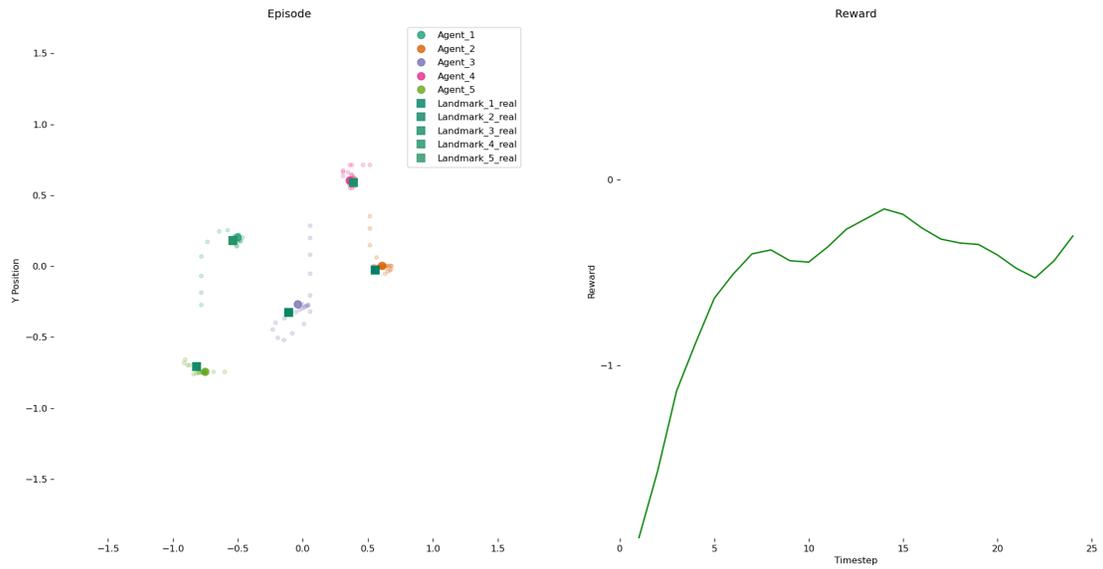

(c) 5 Agents, 5 Landmarks

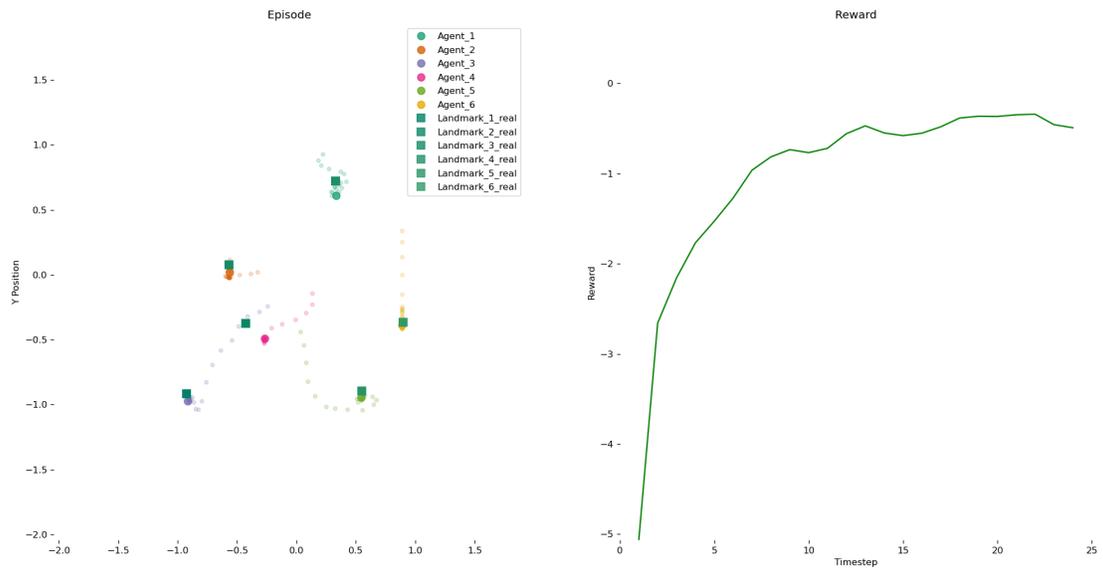

(d) 6 Agents, 6 Landmarks

Figure 4: Some examples of the learned policies in the Spread tasks, using TransfQMix trained in the 4v4 scenario. The smoothed circles represent the trajectories of the agents. The full-filled circles represent their positions at the end of the episode. The green line in the right figures is the evolution of the global reward during the episode.